\documentclass[11pt]{article}

\usepackage[margin=1in]{geometry}
\usepackage{amsmath}
\usepackage{booktabs}
\usepackage{graphicx}
\usepackage{multirow}
\usepackage{url}
\usepackage{xcolor}
\usepackage[hidelinks]{hyperref}
\usepackage{caption}
\usepackage{microtype}

\newcommand{\modelbig}{LexisLM-1B}
\newcommand{\modelsmall}{LexisLM-500M}

\title{\textbf{Compact Greek Language Models from Scratch:\\
Pretraining 500M- and 1B-Parameter Decoder-Only Transformers}}
\author{%
  \begin{tabular}{ccc}
    \small Konstantinos Perifanos\textsuperscript{1} &
    \small Anne Laure Poite &
    \small Giannis Tsilikas\textsuperscript{2} \\
    \small Ilias Katis\textsuperscript{2} &
    \small Stavros Deligiannidis\textsuperscript{2} &
    \small Dionysis Goutsos\textsuperscript{1} \\
    \multicolumn{3}{c}{\small\textsuperscript{1} National and Kapodistrian University of Athens (NKUA)} \\
    \multicolumn{3}{c}{\small\textsuperscript{2} lightsaber.gr}
  \end{tabular}
}
\date{November 2026}

\begin{document}

\maketitle

\begin{abstract}
We present two compact decoder-only language models trained from scratch
exclusively on Greek text: \modelbig{} (1.09B parameters) and \modelsmall{}
(502M parameters). Both use a dedicated 32k byte-level BPE tokenizer and a
22B-token pretraining corpus drawn primarily from Greek Wikipedia and FineWeb2.
Both models complete the 22B-token schedule, with best held-out
perplexities of 9.77 for \modelbig{} and 11.31 for \modelsmall{}.
Evaluation of both models on Greek-translated ARC, MMLU, and TruthfulQA
shows above-chance performance on ARC-Easy, while ARC-Challenge and MMLU
remain near chance.
We also perform supervised fine-tuning of both models on Greek instruction
data, primarily EuroAlpaca, reaching held-out instruction perplexities of
7.12 for \modelsmall{} and 6.37 for \modelbig{}.
The report describes the tokenizer, data preparation, training, and evaluation
methods, providing a foundation for further work on compact Greek language
models.
\end{abstract}

\section{Introduction}
\label{sec:intro}

Recent open Greek generative models, notably Meltemi and Krikri, build on
existing multilingual checkpoints through continued
pretraining~\cite{voukoutis2024meltemi,voukoutis2025krikri}. This report
investigates a different approach: pretraining compact, Greek-first language
models from random initialization, without inheriting weights from an
existing model. Our main contribution is Greek-only pretraining at the
500M and 1B parameter scales, using a
dedicated Greek tokenizer, a 22B-token corpus, and a shared training recipe.
We also report supervised fine-tuning (SFT) of both models, based primarily
on the Greek portion of EuroAlpaca~\cite{sant2026euroalpaca}.

The release consists of model weights. The Hugging Face page for the model
family is:
\begin{center}
\url{https://huggingface.co/ComputationalLinguisticsLab}.
\end{center}

The contributions of this report are:
\begin{itemize}
  \item A 32k byte-level BPE tokenizer trained on a weighted Greek
        Wikipedia/FineWeb2 sample, with a detailed fertility and coverage
        analysis (Section~\ref{sec:tokenizer}).
  \item A description of the 22.0B-token Greek pretraining corpus and its
        preparation (Section~\ref{sec:data}).
  \item From-scratch pretraining of two decoder-only models,
        \modelsmall{} (502M) and \modelbig{} (1.09B), using a shared training
        recipe. We report completed training and held-out language-modeling
        results and Greek-translated ARC, MMLU, and TruthfulQA benchmark
        evaluations for both models
        (Sections~\ref{sec:models}--\ref{sec:results}).
  \item SFT of both model sizes using a shared EuroAlpaca-Greek-centered
        mixture and an append-only chat-token extension of the
        tokenizer (Section~\ref{sec:sft}).
\end{itemize}

\section{Related Work}
\label{sec:related-work}

\paragraph{Greek language models.}
Earlier monolingual work includes GREEK-BERT~\cite{koutsikakis2020greekbert},
an encoder model evaluated on part-of-speech tagging, named entity
recognition, and natural language inference. This establishes important
prior work in Greek language-model pretraining, but differs from the
decoder-only, autoregressive models studied here.

Greek generative models also predate recent multi-billion-parameter
releases. The \nolinkurl{nikokons/gpt2-greek} model
card\footnote{\url{https://huggingface.co/nikokons/gpt2-greek}}
explicitly reports training a small GPT-2 architecture from scratch on
almost 5GB of Greek text, primarily Wikipedia, organized into approximately
3.7 million five-sentence samples. Training uses a 512-token context and
20 epochs. Its published configuration specifies 12 layers, a hidden
dimension of 768, and a 22{,}000-entry vocabulary, corresponding to
approximately 102M parameters under standard GPT-2 embedding tying.
In contrast, \nolinkurl{lighteternal/gpt2-finetuned-greek}
adapts an English GPT-2 checkpoint to Greek~\cite{lighteternalgreekgpt2}.
Thus, neither compact Greek generation nor from-scratch Greek pretraining
is new in itself.

Table~\ref{tab:greek-scratch-comparison} compares the earlier from-scratch
model with our approach. Our 502M and 1.09B models are approximately five
and eleven times larger, respectively, and use an eightfold longer context.
We also replace the GPT-2 architecture with LLaMA-style components and
train on a 22B-token Wikipedia/FineWeb2 corpus without planned data
repetition. The earlier model card does not report either the tokenized
corpus size or the total training tokens processed. Consequently, its
5GB corpus and 20-epoch schedule cannot be converted into a reliable
token-budget ratio from the available documentation.

\begin{table}[htbp]
\centering
\small
\begin{tabular}{lccc}
\toprule
\textbf{Property} & \textbf{Greek GPT-2} & \textbf{\modelsmall} & \textbf{\modelbig} \\
\midrule
Initialization       & Random & Random & Random \\
Architecture         & GPT-2 & LLaMA-style & LLaMA-style \\
Parameters           & $\approx$102M & 502M & 1.09B \\
Context (tokens)     & 512 & 4{,}096 & 4{,}096 \\
Base vocabulary      & 22{,}000 & 32{,}000 & 32{,}000 \\
Corpus tokens        & Not reported & 22B & 22B \\
Tokens processed     & Not reported & 22B & 22B \\
Training passes      & 20 epochs & Single pass & Single pass \\
\bottomrule
\end{tabular}
\caption{Comparison of Greek models pretrained from scratch. Greek GPT-2
details are from its model card and published configuration; its parameter
count is a configuration-based estimate. Our models reserve four additional
vocabulary entries for chat tokens. Both runs have completed.}
\label{tab:greek-scratch-comparison}
\end{table}

\paragraph{Continued pretraining for Greek.}
Meltemi-7B~\cite{voukoutis2024meltemi} adapts Mistral-7B through continued
pretraining and extends its tokenizer with Greek tokens.
Krikri-8B~\cite{voukoutis2025krikri} similarly builds on Llama 3.1-8B,
using continued pretraining and a multi-stage post-training pipeline.
These models inherit knowledge and representations from their base
checkpoints. Our models instead learn their parameters from random
initialization on Greek text, using a separately trained tokenizer.
This distinction concerns the origin of the model weights, not whether
the architecture is new: we adopt established LLaMA-style components.

\paragraph{Compact pretraining and scope of this report.}
Beyond Greek, TinyLlama~\cite{zhang2024tinyllama} demonstrates pretraining
at the 1.1B parameter scale while reusing the Llama 2 architecture and
tokenizer. Compact model pretraining is therefore an established research
direction. Our focus is Greek-only pretraining at two compact
scales, with tokenizer analysis, a 22B-token corpus, and an explicit
separation between pretraining and subsequent instruction tuning. We do
not claim the first Greek model trained from scratch or superiority over
larger adapted models. Rather, we examine what can be learned from Greek
data alone at these parameter scales under a shared 22B-token budget.

\section{Tokenizer}
\label{sec:tokenizer}

\subsection{Design and training data}

We train a byte-level BPE tokenizer~\cite{sennrich2016bpe,radford2019gpt2}
with the HuggingFace \texttt{tokenizers} library. Text is normalized to
Unicode NFC; case and accents are preserved (no lowercasing, no accent
stripping), and pre-tokenization is byte-level, so the tokenizer has no
unknown tokens and round-trips arbitrary UTF-8.

The training sample contains \textbf{20 million word occurrences}, mixed by
Unicode letter-word occurrences (not bytes or documents): 4M words (20\%) from
Greek Wikipedia and 16M words (80\%) from the Greek subset of
FineWeb2~\cite{penedo2025fineweb2}. The vocabulary has \textbf{32{,}000
entries}, including four special tokens (\texttt{<unk>}, \texttt{<bos>},
\texttt{<eos>}, \texttt{<pad>}), with a BPE minimum merge frequency of 3.
Table~\ref{tab:tok-config} summarizes the configuration.

\begin{table}[htbp]
\centering
\small
\begin{tabular}{ll}
\toprule
\textbf{Property} & \textbf{Value} \\
\midrule
Algorithm & Byte-level BPE \\
Vocabulary size (base) & 32{,}000 (incl.\ 4 special tokens) \\
Vocabulary size (chat-extended) & 32{,}004 \\
Normalization & NFC; case- and accent-preserving \\
Pre-tokenization & Byte-level, no prefix space \\
Minimum merge frequency & 3 \\
Training sample & 20M words: 20\% Wikipedia / 80\% FineWeb2-Greek \\
Unknown tokens & 0 (byte-level coverage) \\
Special tokens (base) & \texttt{<unk>}, \texttt{<bos>}, \texttt{<eos>}, \texttt{<pad>} \\
Chat tokens (extension) & \texttt{<|system|>}, \texttt{<|user|>}, \texttt{<|assistant|>}, \texttt{<|endofturn|>} \\
\bottomrule
\end{tabular}
\caption{Tokenizer configuration.}
\label{tab:tok-config}
\end{table}

\subsection{Fertility and coverage}

Fertility is the number of token pieces per word; lower is better. Every word
is encoded with one leading space and no special tokens, case-sensitively and
accent-preserving. The held-out sample contains 200{,}000 words with the same
20/80 source mixture (40k Wikipedia / 160k FineWeb2), with Wikipedia articles
split by seeded hash and FineWeb evaluation shards disjoint from training
shards. The primary metrics consider words appearing at least three times in
the sampled training corpus; ``Greek'' means containing at least one Greek
letter. Coverage is the share of held-out Greek word occurrences meeting the
frequency criterion.

\begin{table}[htbp]
\centering
\small
\begin{tabular}{lccc}
\toprule
\textbf{Source} & \textbf{Greek fertility} & \textbf{Single-token} & \textbf{Eligible occurrence} \\
 & \textbf{(occurrence-weighted)} & \textbf{words} & \textbf{coverage} \\
\midrule
Wikipedia        & 1.268 & 81.30\% & 95.66\% \\
FineWeb2         & 1.210 & 84.53\% & 96.89\% \\
20/80 mixture    & \textbf{1.221} & \textbf{83.91\%} & \textbf{96.65\%} \\
\bottomrule
\end{tabular}
\caption{Held-out Greek word fertility of the 32k tokenizer.}
\label{tab:fertility}
\end{table}

Additional statistics beyond Table~\ref{tab:fertility}:
\begin{itemize}
  \item Mixed held-out Greek fertility with equal weight per distinct eligible
        word form (type-weighted) is \textbf{1.931}; the type-weighted
        single-token rate is 34.21\%.
  \item Across eligible Greek word occurrences, the median is one token and
        the 90th and 95th percentiles are two tokens; 4.84\% require at least
        three tokens, with a maximum of eight. For distinct eligible Greek
        word types, the median is two tokens, the 90th and 95th percentiles
        are three, and 21.94\% require at least three tokens (maximum eight).
  \item On the full sampled training lexicon, \textbf{202{,}077} distinct
        Greek word forms meet the frequency threshold; their type-weighted
        fertility is \textbf{2.654} and occurrence-weighted fertility
        \textbf{1.235}. Common words dominate occurrence-weighted averages,
        explaining the gap.
  \item \textbf{1.96\%} of held-out Greek word occurrences are unseen in the
        sampled training lexicon. This is distinct from eligible-occurrence
        coverage: the latter excludes all words with fewer than three
        occurrences in the tokenizer's sampled training corpus.
  \item Across all languages/scripts in the mixed held-out sample,
        occurrence-weighted fertility is \textbf{1.243} with \textbf{82.65\%}
        single-token coverage.
\end{itemize}

These statistics show why both weighting schemes matter. Frequent Greek words
are compactly encoded: the occurrence-weighted 95th percentile is only two
tokens, and more than four fifths of eligible occurrences are single-token.
Rare word forms are split more often, as reflected by the higher type-weighted
fertility and its 21.94\% three-or-more-token share. This is a long-tail
effect rather than poor coverage of common text; byte-level encoding still
ensures that unseen spellings can be represented.

\subsection{Full-text compression and robustness}

On the complete held-out text --- including rare words, punctuation, and
whitespace, with no frequency filter --- the tokenizer produces
\textbf{314{,}629 tokens} for 2{,}400{,}730 UTF-8 bytes (1{,}346{,}577
characters), i.e.\ \textbf{7.630 bytes/token} and \textbf{4.280
characters/token}, or \textbf{233.65 tokens per 1{,}000 characters}. By source,
the tokenizer averages 7.00 bytes/token on Wikipedia and 7.81 on FineWeb2;
the latter's higher rate reflects the different text mix and should not be
read as a direct quality comparison. There are \textbf{0 unknown tokens} and
\textbf{0 round-trip failures} across 7{,}967 evaluated lines, and NFC and
decomposed Greek spellings encode identically. These checks indicate robust
UTF-8 coverage and lossless decoding, while the bytes-per-token figure
provides a text-level compression measure complementary to word fertility.

\subsection{Chat-token extension}

For instruction tuning we extend the tokenizer with four chat-control tokens
(\texttt{<|system|>}, \texttt{<|user|>}, \texttt{<|assistant|>},
\texttt{<|endofturn|>}) at IDs 32{,}000--32{,}003, yielding a vocabulary of
\textbf{32{,}004}. The extension is \emph{append-only}: all 32{,}000 base
IDs, merges, and normalization are preserved, so pretraining token arrays
remain valid without retokenization. Both pretrained models are built
with \texttt{vocab\_size} 32{,}004 so SFT requires no embedding resizing.

\section{Pretraining Data}
\label{sec:data}

The pretraining corpus is tokenized with the base tokenizer into flat
\texttt{uint16} arrays, one \texttt{<eos>} (ID 2) appended per source
document. The job targets 22B training tokens and stops at the end of the
chunk that crosses the target, with no oversampling.

\begin{table}[htbp]
\centering
\small
\begin{tabular}{lrr}
\toprule
\textbf{Split} & \textbf{Tokens} & \textbf{Documents} \\
\midrule
Train       & 22{,}002{,}557{,}667 & 34{,}079{,}145 \\
Validation  & 32{,}218{,}799       & 46{,}477 \\
\bottomrule
\end{tabular}
\caption{Pretraining corpus statistics (EOS tokens included).}
\label{tab:corpus}
\end{table}

\paragraph{Sources and splits.} All extracted Greek Wikipedia is included,
using the same seeded article-content-hash split as the tokenizer sample (5\%
held out for validation). The FineWeb2 Greek subset supplies
the remainder: its official \texttt{test} split goes to validation, and its
official \texttt{train} split is streamed sequentially until the token target
is reached. 

\paragraph{Data ordering.} Training uses a single pass over the corpus,
with 4{,}096-token sequences shuffled using a fixed seed. Padding in the
final batch is excluded from the training loss.

\section{Models}
\label{sec:models}

\subsection{Architecture}

Both models are decoder-only Transformers following the LLaMA
recipe~\cite{touvron2023llama}: pre-norm RMSNorm~\cite{zhang2019rmsnorm},
rotary position embeddings (RoPE, $\theta = 10^4$)~\cite{su2021roformer},
SwiGLU feed-forward blocks~\cite{shazeer2020glu} with hidden ratio $8/3$
(rounded to hardware-friendly sizes), no bias terms anywhere, and input/output
embedding tying. Attention uses PyTorch scaled-dot-product attention with a
causal mask. Table~\ref{tab:arch} lists the two configurations.

\begin{table}[htbp]
\centering
\small
\begin{tabular}{lcc}
\toprule
 & \textbf{\modelsmall} & \textbf{\modelbig} \\
\midrule
Parameters (total, tied)   & 502{,}193{,}664 & 1{,}093{,}232{,}640 \\
Layers                     & 16    & 20 \\
Model dimension $d$        & 1{,}536 & 2{,}048 \\
Attention heads            & 16    & 16 \\
Head dimension             & 96    & 128 \\
SwiGLU hidden dimension    & 4{,}096 & 5{,}632 \\
Context length             & 4{,}096 & 4{,}096 \\
Vocabulary                 & 32{,}004 & 32{,}004 \\
Positional encoding        & RoPE ($\theta{=}10^4$) & RoPE ($\theta{=}10^4$) \\
Normalization              & RMSNorm ($\epsilon{=}10^{-5}$) & RMSNorm ($\epsilon{=}10^{-5}$) \\
Biases / Dropout           & none / 0.0 & none / 0.0 \\
Embedding tying            & yes & yes \\
\bottomrule
\end{tabular}
\caption{Model architectures.}
\label{tab:arch}
\end{table}

\paragraph{Architectural choices.} The 4{,}096-token context is matched to the
shuffled-block data ordering for single-pass traversal of token blocks. We tie
the input and output embeddings to reduce the parameter count: with a 32k
vocabulary, tying saves $\sim$66M (\modelbig) or $\sim$49M (\modelsmall)
parameters compared with untied embeddings. The $8/3$ SwiGLU ratio keeps
parameter count comparable to a $4\times$ GELU MLP while benefiting from
gating. We keep a modest RoPE base of $10^4$ since we do not target context
extension. Vocabulary is set to 32{,}004 from the start so that the chat
extension (Section~\ref{sec:tokenizer}) requires no architecture change.

\subsection{Pretraining setup}

Both models use the same optimizer recipe (Table~\ref{tab:pretrain-hparams})
and the same token budget: $335{,}733$ steps $\times$ $65{,}536$ tokens/step
$\approx22.0$B tokens, approximately one pass over the corpus. Training runs in
\texttt{bfloat16} on a single CUDA device. Validation loss, perplexity, and bits-per-byte (using
the pretraining tokenizer for byte counts) are measured throughout.

\paragraph{Token-budget rationale.}
The 22B-token budget is motivated by the compute-optimal scaling analysis
of Hoffmann et al.~\cite{hoffmann2022chinchilla}, which finds that model
size and training tokens should increase in approximately equal
proportions as training compute grows. A commonly used approximation to
the Chinchilla allocation is $D \approx 20N$, where $N$ is the parameter
count and $D$ is the number of training tokens. For \modelbig{}, this gives
\[
  D \approx 20 \times 1.093\text{B} \approx 21.9\text{B tokens},
\]
motivating a rounded target of 22B tokens, or approximately 20.1 tokens
per parameter. This is a budget-selection heuristic, not a demonstrated
compute optimum for our Greek corpus, tokenizer, or architecture.

For \modelsmall{}, the same heuristic would suggest approximately 10.0B
tokens. We retain the shared 22B-token target, corresponding to
approximately 43.8 tokens per parameter for the completed run, to
compare model sizes under a matched data budget and training schedule.
The smaller model is therefore trained beyond the heuristic allocation,
rather than at a separately optimized compute budget. Training compact
models on more tokens can also serve inference-oriented
objectives~\cite{touvron2023llama}, although we do not optimize total
training and inference cost here. 

\begin{table}[htbp]
\centering
\small
\begin{tabular}{lcc}
\toprule
 & \textbf{\modelsmall} & \textbf{\modelbig} \\
\midrule
Optimizer                & \multicolumn{2}{c}{AdamW ($\beta_1{=}0.9$, $\beta_2{=}0.95$)} \\
Peak learning rate       & \multicolumn{2}{c}{$3\times10^{-4}$} \\
Schedule                 & \multicolumn{2}{c}{cosine to $3\times10^{-5}$, 2{,}000 warmup steps} \\
Weight decay             & \multicolumn{2}{c}{0.1} \\
Gradient clipping        & \multicolumn{2}{c}{1.0} \\
Precision                & \multicolumn{2}{c}{bfloat16} \\
Tokens per step          & \multicolumn{2}{c}{$16 \times 4{,}096 = 65{,}536$} \\
Micro-batch $\times$ accumulation & $8 \times 2$ & $2 \times 8$ \\
Max steps                & \multicolumn{2}{c}{335{,}733} \\
Total tokens             & \multicolumn{2}{c}{22.0B} \\
Data order               & \multicolumn{2}{c}{seeded block shuffle, single pass (seed 1337)} \\
\bottomrule
\end{tabular}
\caption{Pretraining hyperparameters.}
\label{tab:pretrain-hparams}
\end{table}

\section{Pretraining Results}
\label{sec:results}

\subsection{Language-modeling quality}

Both models completed the full training schedule, consuming
approximately 22.0B tokens each. Table~\ref{tab:lm-results} reports held-out
validation metrics for their best evaluated checkpoints.

Both models are evaluated on the shared Wikipedia/FineWeb2-Greek validation
split.

\begin{table}[htbp]
\centering
\small
\begin{tabular}{lcccc}
\toprule
\textbf{Model} & \textbf{Tokens seen} & \textbf{Val.\ loss} & \textbf{Val.\ PPL} & \textbf{Bits/byte} \\
\midrule
\modelsmall{} (best) & 21.8B & 2.426 & 11.31 & 0.466 \\
\modelbig{} (best)   & 22.0B & \textbf{2.280} & \textbf{9.77} & \textbf{0.435} \\
\bottomrule
\end{tabular}
\caption{Held-out validation metrics on the pretraining validation split
(Wikipedia + FineWeb2-Greek). Token positions seen refer to the selected checkpoints,
not the completed runs; both runs consumed 22.0B tokens. \modelsmall{}
metrics are from step 332{,}000 and \modelbig{} metrics from step
335{,}000.}
\label{tab:lm-results}
\end{table}

Under the shared 22B-token training budget, the best evaluated
\modelbig{} checkpoint has a validation loss of 2.280, compared with
2.426 for \modelsmall{}. This suggests a language-modeling advantage for
the larger model under the shared training data order and schedule.

\subsection{Greek benchmark evaluation}

{\sloppy
We evaluate on Greek-translated benchmarks published by
ILSP~\cite{voukoutis2024meltemi}: ARC-Easy and ARC-Challenge
(\texttt{ilsp/arc\_greek}~\cite{ilsp2024arcgreek}), MMLU
(\texttt{ilsp/mmlu\_greek}~\cite{ilsp2024mmlugreek}), and
TruthfulQA multiple-choice
(\texttt{ilsp/truthful\_qa\_greek}~\cite{ilsp2024truthfulqagreek}).
Scoring uses answer log-likelihood with character-length normalization for
ARC (reported as normalized accuracy), standard accuracy for MMLU, and
MC1/MC2 for TruthfulQA. The few-shot protocol follows common practice:
ARC 25-shot, MMLU 5-shot, TruthfulQA 0-shot.\par}

Both models are evaluated on all four tasks using their best
training-loop validation checkpoints, selected near the end of their
completed schedules. Demonstration selection uses seed 1337, with the same
prompts, dataset revisions, and scoring procedure for both models.
ARC demonstrations come from its training splits, which also supply
the small pretraining component described in Section~\ref{sec:data}.
We report MMLU accuracy aggregated over questions.

\begin{table}[htbp]
\centering
\small
\begin{tabular}{lccccc}
\toprule
\textbf{Model / checkpoint} & \textbf{ARC-C} & \textbf{ARC-E} & \textbf{MMLU} & \textbf{TQA MC1} & \textbf{TQA MC2} \\
 & (norm.) & (raw / norm.) & & & \\
\midrule
\modelsmall{}, few-shot, 332k & 25.3 & 44.4 / 41.9 & 25.3 & 24.7 & 42.1 \\
\modelbig{}, few-shot, 335k (best) & 26.7 & 46.3 / 44.6 & 24.5 & 23.9 & 41.4 \\
\midrule
Random baseline          & 25.0 & 25.0 & 25.0 & $\sim$22.7 & --- \\
\bottomrule
\end{tabular}
\caption{Greek benchmark accuracy (\%). Few-shot: ARC
25-shot, MMLU 5-shot, TruthfulQA 0-shot. Question counts: ARC-C 1{,}168,
ARC-E 2{,}376, MMLU 14{,}042, TruthfulQA 817. MMLU scores are accuracy
aggregated over questions. Both few-shot evaluations use the requested
number of demonstrations for every question, with no question truncation.}
\label{tab:benchmarks}
\end{table}

\paragraph{Discussion.} Both models exceed chance on ARC-Easy in raw accuracy,
while ARC-Challenge and MMLU remain near the nominal 25\% chance
baseline.

The best \modelsmall{} checkpoint (step 332{,}000) achieves 25.3\% normalized accuracy
on ARC-Challenge with 25-shot prompting (20.3\% raw accuracy) and 25.3\%
accuracy on MMLU with 5-shot prompting, both close to the nominal 25\%
chance baseline. Zero-shot TruthfulQA scores are 24.7\% MC1 and 42.1\%
MC2. On ARC-Easy, \modelsmall{} scores 44.4\% raw and 41.9\% normalized
accuracy, compared with 46.3\% and 44.6\% for \modelbig{} under the same
few-shot protocol. The larger model also scores higher on ARC-Challenge,
whereas the smaller model scores higher on MMLU and TruthfulQA.

These results should be interpreted in light of both model capacity and
pretraining history. At 502M and 1.09B parameters, our models have
substantially less capacity than Meltemi-7B and
Krikri-8B~\cite{voukoutis2024meltemi,voukoutis2025krikri} to represent
the broad factual knowledge and reasoning patterns tested by these
benchmarks. Moreover, Meltemi and Krikri continue pretraining from
Mistral-7B and Llama 3.1-8B, respectively, inheriting representations
and knowledge from large-scale multilingual pretraining. Our models
instead learn from random initialization using only 22B Greek tokens.
 
Nevertheless, the lower held-out perplexity and higher ARC-Easy and
ARC-Challenge scores of \modelbig{} provide preliminary encouragement
for exploring larger Greek models trained from scratch. The evidence
is promising as motivation for further experiments.

\section{Supervised Fine-Tuning on EuroAlpaca}
\label{sec:sft}

\subsection{EuroAlpaca-Greek}

EuroAlpaca~\cite{sant2026euroalpaca} is a task-preserving localization of the
Alpaca instruction dataset~\cite{taori2023alpaca} into European languages,
distributed under CC BY-NC 4.0. We use the Greek configuration:
\textbf{51{,}760} source rows of \texttt{(instruction, input, output)}
triples.

Preparation performs validation and prompt-level deduplication:
\begin{itemize}
  \item 1 row dropped for an empty instruction/output field;
  \item exact-duplicate prompts removed under an NFC, whitespace-collapsed,
        case-sensitive key over \texttt{(instruction, input)}, keeping the
        lowest source index: \textbf{25 rows removed}, of which 19 had
        conflicting outputs for the same prompt;
  \item final dataset: \textbf{51{,}734 examples}, each tokenized and stored
        with per-example loss boundaries.
\end{itemize}

The SFT mixture is dominated by EuroAlpaca (80.8\%) with two small auxiliary
Greek sets for conversational coverage (Table~\ref{tab:sft-data}).

\begin{table}[htbp]
\centering
\small
\begin{tabular}{lrr}
\toprule
\textbf{Dataset} & \textbf{Examples} & \textbf{Share} \\
\midrule
EuroAlpaca-Greek (dedup.)        & 51{,}734 & 80.8\% \\
Greek chat SFT                   & 11{,}675 & 18.2\% \\
Aya-Greek~\cite{singh2024aya}    & 609      & 1.0\% \\
\midrule
Total                            & 64{,}018 & 100\% \\
\quad train / validation (5\%)   & 60{,}818 / 3{,}200 & \\
\bottomrule
\end{tabular}
\caption{SFT data mixture.}
\label{tab:sft-data}
\end{table}

\subsection{Chat template and loss masking}

Examples are rendered with the four chat-control tokens added in
Section~\ref{sec:tokenizer}:

\begin{center}
\small
\texttt{<|user|>}\,\textit{instruction}\,[\texttt{\textbackslash n\textbackslash n}\,\textit{input}]\,\texttt{<|endofturn|>}\,\texttt{<|assistant|>}\,\textit{output}\,\texttt{<|endofturn|>}
\end{center}

Sequences start from an \texttt{<eos>} prefix token. The loss is computed only
on assistant content and the closing turn token; user/system tokens and the
assistant role prefix are masked out. Sequences are not packed, and inputs
longer than 4{,}096 tokens are left-truncated (preserving the answer).

\subsection{SFT setup}

SFT initializes each model from its best pretraining checkpoint:
step 332{,}000 for \modelsmall{} and step 335{,}000 for \modelbig{}.
Both runs use the same instruction mixture, seeded train/validation split
(seed 1337), chat template, and hyperparameters
(Table~\ref{tab:sft-hparams}). Both models reserve space for the four
chat tokens during pretraining, so SFT requires no embedding resizing.

\begin{table}[htbp]
\centering
\small
\begin{tabular}{ll}
\toprule
\textbf{Setting} & \textbf{Value} \\
\midrule
Initialization        & Each model's best pretraining checkpoint \\
Steps                 & 3{,}802 ($\approx$1 epoch over 60{,}818 examples) \\
Batch                 & 2 $\times$ 8 accumulation = 16 sequences/step \\
Learning rate         & $2\times10^{-5}$, cosine to $2\times10^{-6}$, 100 warmup steps \\
Weight decay          & 0.0 \\
Optimizer             & AdamW ($\beta_1{=}0.9$, $\beta_2{=}0.95$), clip 1.0 \\
Precision             & bfloat16 \\
Packing / truncation  & none / left at 4{,}096 \\
\bottomrule
\end{tabular}
\caption{Shared SFT hyperparameters for both model sizes.}
\label{tab:sft-hparams}
\end{table}

\subsection{SFT results}

Both models complete 3{,}802 SFT steps. On the shared held-out
3{,}200-example instruction split, both achieve their lowest validation
loss at the final evaluation (Table~\ref{tab:sft-validation}).
The larger model has lower instruction perplexity and higher next-token
accuracy on supervised positions. These instruction-validation metrics
use assistant-only loss masking and are not directly comparable to the
pretraining perplexities in Table~\ref{tab:lm-results}. The observed
validation minima do not establish whether additional epochs would
improve performance or lead to overfitting.

\begin{table}[htbp]
\centering
\small
\begin{tabular}{lrrrr}
\toprule
\textbf{Model} & \textbf{SFT step} & \textbf{Val.\ loss} & \textbf{Val.\ PPL} & \textbf{Token accuracy} \\
\midrule
\modelsmall{} + SFT & 3{,}802 & 1.963 & 7.12 & 56.4\% \\
\modelbig{} + SFT   & 3{,}802 & 1.852 & 6.37 & 58.1\% \\
\bottomrule
\end{tabular}
\caption{Instruction-validation results on the same 3{,}200 examples.
Loss, perplexity, and next-token accuracy are computed over supervised
assistant content and closing turn tokens.}
\label{tab:sft-validation}
\end{table}

\begin{table}[htbp]
\centering
\small
\begin{tabular}{lccccc}
\toprule
\textbf{Model} & \textbf{ARC-C} & \textbf{ARC-E} & \textbf{MMLU} & \textbf{TQA MC1} & \textbf{TQA MC2} \\
 & (norm.) & (raw / norm.) & & & \\
\midrule
\modelsmall{} base (few-shot, 332k) & 25.3 & 44.4 / 41.9 & 25.3 & 24.7 & 42.1 \\
\modelsmall{} + SFT (few-shot)      & 25.3 & 43.7 / 41.6 & 26.0 & 23.1 & 41.3 \\
\midrule
\modelbig{} base (few-shot, 335k) & 26.7 & 46.3 / 44.6 & 24.5 & 23.9 & 41.4 \\
\modelbig{} + SFT (few-shot)      & 27.9 & 46.1 / 44.1 & 24.7 & 22.5 & 40.9 \\
\bottomrule
\end{tabular}
\caption{Greek benchmarks before and after SFT (\%). Base rows use the
same checkpoints used to initialize SFT: step 332{,}000 for \modelsmall{}
and step 335{,}000 for \modelbig{}. The protocol is the same as
Table~\ref{tab:benchmarks} (ARC 25-shot, MMLU 5-shot, TruthfulQA 0-shot).
Both SFT models are evaluated at step 3{,}802.}
\label{tab:sft-benchmarks}
\end{table}

For \modelsmall{}, SFT increases MMLU accuracy from 25.3\% to 26.0\%.
ARC-Challenge normalized accuracy remains 25.3\% at the reported precision.
ARC-Easy raw and normalized accuracy decrease from 44.4\% to 43.7\% and
41.9\% to 41.6\%, respectively. TruthfulQA MC1 decreases from 24.7\% to
23.1\%, and MC2 from 42.1\% to 41.3\%.

For \modelbig{}, SFT increases ARC-Challenge
normalized accuracy from 26.7\% to 27.9\% and MMLU accuracy from 24.5\%
to 24.7\%. ARC-Easy raw and normalized accuracy decrease slightly
(46.3\% to 46.1\% and 44.6\% to 44.1\%, respectively), as do TruthfulQA
MC1 (23.9\% to 22.5\%) and MC2 (41.4\% to 40.9\%).
These mixed changes do not demonstrate uniform benchmark improvement or
preservation of all base-model capabilitie. The multiple-choice evaluations do not directly measure
instruction following. Qualitatively, the SFT model follows the chat
template in observations of \modelbig{}, terminating turns with
\texttt{<|endofturn|>}; a systematic
instruction-following evaluation remains future work.

\section{Limitations}
\label{sec:limitations}

\begin{itemize}
  \item \textbf{Fixed training budget.} We use a 22B-token, approximately
        single-pass pretraining schedule. We do not investigate whether
        additional fresh data or repeated exposure to the existing corpus
        would improve performance.
  \item \textbf{Greek-only pretraining.} We do not investigate multilingual
        transfer; knowledge-heavy benchmarks such as MMLU remain near chance
        at this scale.
  \item \textbf{Benchmark translation effects.} Greek-translated ARC, MMLU,
        and TruthfulQA inherit translation artifacts; absolute numbers are not
        directly comparable to English evaluations.
  \item \textbf{SFT licensing.} EuroAlpaca is CC BY-NC 4.0, restricting the
        SFT checkpoints to non-commercial use.
  \item \textbf{No preference tuning.} We report SFT only; no RLHF/DPO stage
        is included, and safety behavior has not been tuned or evaluated.
\end{itemize}

\section{Conclusion}
\label{sec:conclusion}

We described the training and evaluation of compact Greek language
models: a 32k Greek BPE tokenizer with low occurrence-weighted fertility
(1.221 tokens/word and 83.9\% single-token encoding among eligible held-out
Greek word occurrences), a 22B-token corpus, two
decoder-only models at 500M and 1B parameters, each pretrained on 22B tokens,
and an SFT stage centered on
EuroAlpaca-Greek using an append-only chat-token extension. Both models
achieve above-chance raw accuracy on ARC-Easy (44.4\% for \modelsmall{}
and 46.3\% for \modelbig{} with 25-shot prompting), subject to their ARC
training exposure. Their best held-out perplexities are 11.31 and 9.77,
respectively. Both models undergo instruction tuning on the same Greek
mixture, attaining instruction-validation perplexities of 7.12 and 6.37.
Benchmark changes after SFT are mixed rather than uniformly positive at
either model size.

The improvement in held-out perplexity and ARC scores from 500M to 1B
parameters offers a cautiously promising starting point for larger
Greek-only models trained from scratch. However, the absence of gains
on MMLU and TruthfulQA, together with ARC training exposure, means that
these results do not establish a general benchmark scaling trend or
predict parity with larger adapted models such as Meltemi and Krikri.
A natural next step is to evaluate larger models jointly with increased
token budgets and broader, high-quality data, using controlled
comparisons and benchmarks without training exposure. Evaluating fresh
versus repeated data and adding preference optimization are further
directions.

\bibliographystyle{unsrt}
\begingroup
\raggedright
\bibliography{references}

@misc{sant2026euroalpaca,
  title        = {EuroAlpaca: Task-Preserving Localisation of Instruction Data for European Languages},
  author       = {Sant, Aleix and others},
  year         = {2026},
  howpublished = {\url{https://github.com/Telefonica-Scientific-Research/EuroAlpaca}},
  note         = {Dataset: \url{https://huggingface.co/datasets/Telefonica/EuroAlpaca}}
}

@inproceedings{sennrich2016bpe,
  title     = {Neural Machine Translation of Rare Words with Subword Units},
  author    = {Sennrich, Rico and Haddow, Barry and Birch, Alexandra},
  booktitle = {Proceedings of the 54th Annual Meeting of the Association for Computational Linguistics},
  year      = {2016}
}

@article{radford2019gpt2,
  title   = {Language Models are Unsupervised Multitask Learners},
  author  = {Radford, Alec and Wu, Jeffrey and Child, Rewon and Luan, David and Amodei, Dario and Sutskever, Ilya},
  journal = {OpenAI Technical Report},
  year    = {2019}
}

@misc{touvron2023llama,
  title         = {LLaMA: Open and Efficient Foundation Language Models},
  author        = {Touvron, Hugo and Lavril, Thibaut and Izacard, Gautier and others},
  year          = {2023},
  eprint        = {2302.13971},
  archivePrefix = {arXiv}
}

@article{hoffmann2022chinchilla,
  title   = {Training Compute-Optimal Large Language Models},
  author  = {Hoffmann, Jordan and Borgeaud, Sebastian and Mensch, Arthur and Buchatskaya, Elena and Cai, Trevor and others},
  journal = {arXiv preprint arXiv:2203.15556},
  year    = {2022}
}

@inproceedings{zhang2019rmsnorm,
  title     = {Root Mean Square Layer Normalization},
  author    = {Zhang, Biao and Sennrich, Rico},
  booktitle = {Advances in Neural Information Processing Systems},
  year      = {2019}
}

@article{su2021roformer,
  title   = {RoFormer: Enhanced Transformer with Rotary Position Embedding},
  author  = {Su, Jianlin and Lu, Yu and Pan, Shengfeng and Murtadha, Ahmed and Wen, Bo and Liu, Yunfeng},
  journal = {arXiv preprint arXiv:2104.09864},
  year    = {2021}
}

@article{shazeer2020glu,
  title   = {GLU Variants Improve Transformer},
  author  = {Shazeer, Noam},
  journal = {arXiv preprint arXiv:2002.05202},
  year    = {2020}
}

@misc{taori2023alpaca,
  title        = {Stanford Alpaca: An Instruction-following LLaMA model},
  author       = {Taori, Rohan and Gulrajani, Ishaan and Zhang, Tianyi and Dubois, Yann and Li, Xuechen and Guestrin, Carlos and Liang, Percy and Hashimoto, Tatsunori B.},
  year         = {2023},
  howpublished = {\url{https://github.com/tatsu-lab/stanford_alpaca}}
}

@article{voukoutis2024meltemi,
  title   = {Meltemi: The first open Large Language Model for Greek},
  author  = {Voukoutis, Leon and Roussis, Dimitris and Paraskevopoulos, Georgios and Sofianopoulos, Sokratis and Prokopidis, Prokopis and Papavasileiou, Vassilis and Katsamanis, Athanasios and Piperidis, Stelios and Katsouros, Vassilis},
  journal = {arXiv preprint arXiv:2407.20743},
  year    = {2024}
}

@article{voukoutis2025krikri,
  title   = {{Krikri}: Advancing Open Large Language Models for Greek},
  author  = {Voukoutis, Leon and Paraskevopoulos, Georgios and Sofianopoulos, Sokratis and Prokopidis, Prokopis and Papavasileiou, Vassilis and Katsamanis, Athanasios and Piperidis, Stelios and Katsouros, Vassilis},
  journal = {arXiv preprint arXiv:2505.13772},
  year    = {2025}
}

@inproceedings{koutsikakis2020greekbert,
  title     = {{GREEK-BERT}: The Greeks visiting Sesame Street},
  author    = {Koutsikakis, John and Chalkidis, Ilias and Malakasiotis, Prodromos and Androutsopoulos, Ion},
  booktitle = {Proceedings of the 11th Hellenic Conference on Artificial Intelligence},
  year      = {2020},
  doi       = {10.1145/3411408.3411440}
}

@misc{lighteternalgreekgpt2,
  title        = {Greek ({el}) {GPT2} Model: Model Card},
  author       = {{Hellenic Army Academy and Technical University of Crete}},
  howpublished = {\url{https://huggingface.co/lighteternal/gpt2-finetuned-greek}},
  note         = {}
}

@article{zhang2024tinyllama,
  title   = {{TinyLlama}: An Open-Source Small Language Model},
  author  = {Zhang, Peiyuan and Zeng, Guangtao and Wang, Tianduo and Lu, Wei},
  journal = {arXiv preprint arXiv:2401.02385},
  year    = {2024}
}

@misc{ilsp2024arcgreek,
  title        = {{ARC Greek}},
  author       = {{ILSP/Athena RC}},
  year         = {2024},
  howpublished = {\url{https://huggingface.co/datasets/ilsp/arc_greek}},
  note         = {Machine-translated Greek version of the AI2 ARC dataset. }
}

@misc{ilsp2024mmlugreek,
  title        = {{MMLU Greek}},
  author       = {{ILSP/Athena RC}},
  year         = {2024},
  howpublished = {\url{https://huggingface.co/datasets/ilsp/mmlu_greek}},
  note         = {Machine-translated Greek version of the MMLU dataset. }
}

@misc{ilsp2024truthfulqagreek,
  title        = {{Truthful QA Greek}},
  author       = {{ILSP/Athena RC}},
  year         = {2024},
  howpublished = {\url{https://huggingface.co/datasets/ilsp/truthful_qa_greek}},
  note         = {Machine-translated Greek version of the TruthfulQA dataset. }
}

@article{singh2024aya,
  title   = {Aya Dataset: An Open-Access Collection for Multilingual Instruction Tuning},
  author  = {Singh, Shivalika and others},
  journal = {arXiv preprint arXiv:2402.06619},
  year    = {2024}
}

@article{penedo2025fineweb2,
  title={FineWeb2: One Pipeline to Scale Them All--Adapting Pre-Training Data Processing to Every Language},
  author={Penedo, Guilherme and Kydl{\'\i}{\v{c}}ek, Hynek and Sabol{\v{c}}ec, Vinko and Messmer, Bettina and Foroutan, Negar and Kargaran, Amir Hossein and Raffel, Colin and Jaggi, Martin and Von Werra, Leandro and Wolf, Thomas},
  journal={arXiv preprint arXiv:2506.20920},
  year={2025}
}
\endgroup

\end{document}